\pdfoutput = 1
\documentclass{article}
\usepackage{spconf,amsmath,graphicx}
\usepackage{booktabs}
\usepackage[numbers,sort&compress]{natbib}

\usepackage{ graphicx}
\usepackage{ float}

\title{Kronecker Product Feature Fusion for Convolutional Neural Network in Remote Sensing Scene Classification}
\name{Yinzhu Cheng}
\address{Xi'an Institute of Optics and Precision Mechanics of CAS, China\\
    \emph{cyz19970209@163.com}
    }

\begin{document}
%
\maketitle
\begin{abstract}

Remote Sensing Scene Classification is a challenging and valuable research topic, in which Convolutional Neural Network (CNN) has played a crucial role. CNN can extract hierarchical convolutional features from remote sensing imagery, and Feature Fusion of different layers can enhance CNN's performance. Two successful Feature Fusion methods, Add and Concat, are employed in certain state-of-the-art CNN algorithms. In this paper, we propose a novel Feature Fusion algorithm, which unifies the aforementioned methods using the Kronecker Product (KPFF), and we discuss the Backpropagation procedure associated with this algorithm. To validate the efficacy of the proposed method, a series of experiments are designed and conducted. The results demonstrate its effectiveness of enhancing CNN's accuracy in Remote sensing scene classification.
\end{abstract}
\section{Introduction}
\label{sec:intro}

\let\thefootnote\relax
\footnotetext{}

Remote Sensing Scene Classification is one of the most significant research topics of remote sensing, playing a crucial role in urban planning, precision agriculture, mineral exploration, and environmental monitoring \cite{review} \cite{review2} \cite{review3}. To address the rich spatial and structural patterns in remote sensing images, various methods for extracting semantic information have been introduced, such as Scale-Invariant Feature Transform (SIFT) \cite{SIFT}, Histogram of Oriented Gradient (HOG) \cite{HOG}, Bag-of-Words Model (BOW) \cite{BOW}, color histogram (CH) \cite{CH}, and Gray-level Co-occurrence Matrix (GLCM) \cite{GLCM}. 

In recent years, with the widespread application of deep learning methods in the field of remote sensing, Convolutional Neural Network (CNN) has excelled in extracting features of Remote Sensing Scene \cite{CNN1} \cite{CNN2}. Various CNN architectures, including AlexNet \cite{Alex}, VGGNet \cite{VGG}, ResNet \cite{Res}, and Inception \cite{Inception}, have been introduced into remote sensing scene classification, achieving superior performance compared to various traditional methods. In comparison to other methods mentioned earlier, CNN represents an end-to-end approach, where feature extraction and classifier design occur simultaneously \cite{CNN3}. Similar to the human visual system, the shallow layers of CNN focus on contour, edge, texture, and other low-level semantics \cite{CNN4}, while the deeper layers concentrate on high-level semantic features \cite{CNN5}.

Since different layers of CNN can explore the inherent structures and essential information of remote sensing images at different levels, the fusion of features between different layers is beneficial for enhancing the network's performance \cite{FF}. Researchers conducted experiments to study the accuracy of remote sensing scene classification when features from different layers were added or concatenated in CNN \cite{Ma} \cite{Mei} \cite{Xu}. However, the experimental results did not provide a definitive conclusion about which method, Add or Concat, is superior. On the other hand, under certain inappropriate layer combinations, the classification accuracy may even decrease.

To address these issues and enable the network to autonomously learn the importance of each feature, a feature fusion method based on the Kronecker product is proposed. The main contributions of this paper are as follows:

(1) Introducing the classic matrix operation - the Kronecker product into feature fusion in neural networks and unifying the Add and Concat methods. It can be proven that in specific cases, the proposed method degenerates into the classic Add or Concat methods, theoretically not performing weaker than the aforementioned two methods. Through this method, the network can learn a more suitable fusion strategy.

(2) Introducing learnable weight parameters that allow the network to autonomously learn the importance of different features. During the network optimization process, more important features tend to learn larger weight parameters, while the remaining features tend to learn smaller parameters. This is advantageous for addressing the previously mentioned issue of a decrease in classification accuracy after feature fusion.

(3) Conducting basic theoretical analysis of the algorithm, discussing the backpropagation process, and calculating the time complexity. Theoretical analysis indicates that the proposed method has acceptable time complexity in both the forward and backward propagation processes in neural networks.

The rest of this paper is structured as follows. Section 2 provides an overview and summary of the relevant studies on CNN and feature fusion. Section 3 outlines our methodology. In Section 4, we introduce the experimental data and settings, and provide a detailed analysis of the experimental results. Finally, Section 5 contains our conclusion.

\section{Preliminaries}
\label{sec:pre}
\subsection{Feature Fusion}
In neural networks, the two most common feature fusion methods are Add and Concat, which involve either adding or concatenating different features. While there are other feature fusion methods, such as canonical correlation analysis, outer product, Hadamard product, and so on, these methods have not been widely adopted. Add and Concat remain the two most widely used methods in neural networks, and both methods help alleviate the issues of gradient explosion and gradient vanishing.

Let us denote them as $\textbf{x}_1, \textbf{x}_2,...,\textbf{x}_n$ (column vectors), and then the outcome of Add is:
\begin{equation}
	\textbf{y} =	\sum_{i=1}^{n} 	\textbf{x}_i
\end{equation}

While the outcome of Concat is:
\begin{equation}
	\textbf{y} = ||_{i=1}^n	\textbf{x}_i
\end{equation}

\begin{figure}
	\centering
	\includegraphics[width=80mm]{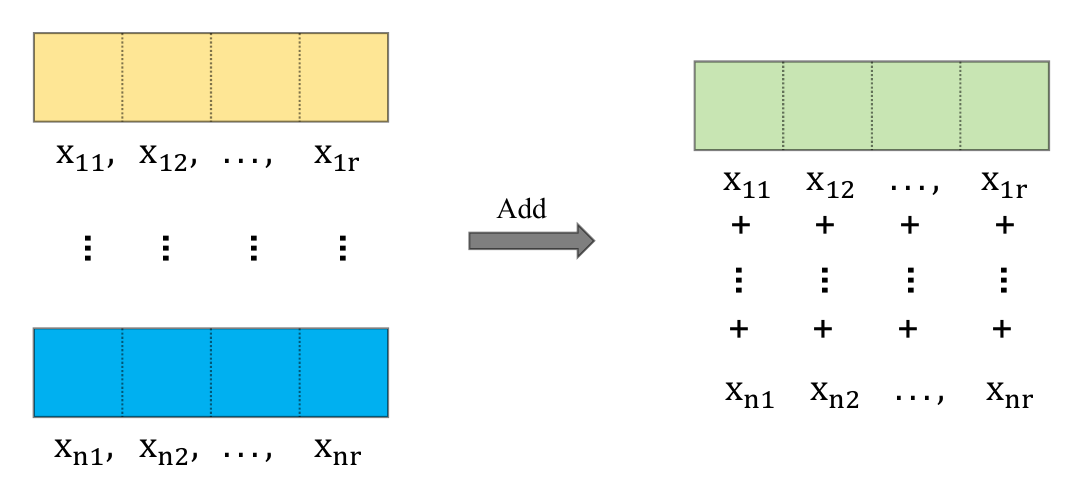}
	\caption{Add Feature Fusion}
	\label{fig1:add}
\end{figure}

\begin{figure}
	\centering
	\includegraphics[width=80mm]{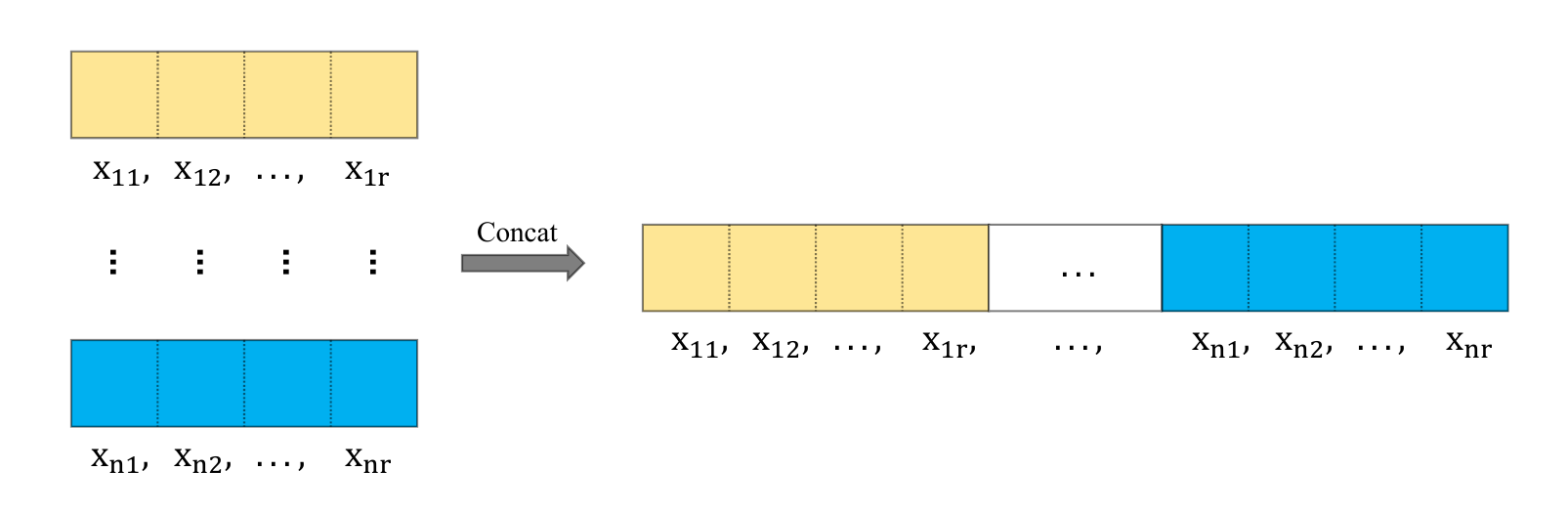}
	\caption{Concat Feature Fusion}
	\label{fig2:concat}
\end{figure}
Add involves element-wise addition of features, assuming a priori that the semantic features of corresponding channels are similar. On the other hand, Concat retains all the information of the feature vectors, allowing the network to autonomously learn a fusion method. Comparative experiments in the literature indicate that these two methods have their respective advantages under different datasets and network structures.

\subsection{Convolutional Neural Network (CNN)}
CNN consists of multiple convolutional layers. Each convolutional layer generates multiple feature maps through convolution operations, thereby extracting image features. If the $l$-th layer has $d_l$ feature maps, and the convolutional kernel has lengths and widths of $2\delta+1$ and $2\gamma+1$, respectively. The value at position $(x, y)$ in the $j$-th feature map of the $i$-th layer, denoted as $v^{x,y}_{i,j}$, through the convolutional kernel with weights of $w^{\sigma,\rho}_{i,j,\tau}$ and bias $b_{i,j}$, and then mapped through the activation function $F$ (such as ReLU, Sigmoid or LeakyReLU), can be expressed as:

\begin{equation}
	v^{x,y}_{i,j} = F(\sum_{\tau=1}^{d_{l-1}} \sum_{\rho = -\gamma}^{\gamma} \sum_{\sigma = -\delta}^{\delta}w^{\sigma,\rho}_{i,j,\tau} v^{x+\sigma,y+\rho}_{i-1,\tau}+ b_{i,j})
\end{equation}

In convolutional neural networks, each feature map can be regarded as a feature of the original image. Feature fusion in such networks allows for the comprehensive utilization of this information, contributing to the enhancement of the CNN's performance. In ResNet and Transformer, the fusion method of Add is employed, while in UNet, Feature Pyramid Network (FPN), and YOLO, the fusion method of Concat is adopted. These network architectures play crucial roles in their respective domains, and their performance often surpasses that of networks without feature fusion. This underscores the importance of feature fusion methods in convolutional neural networks.

After several convolutional layers, pooling layers, and fully connected layers, the output of the CNN can be obtained. Various loss functions compare the network's output with the expected output to calculate the network's loss. By optimizing the network parameters using gradient-based optimization algorithms, the network's output can be brought closer to the expected output, thereby achieving the goal of fitting a function to complete certain tasks.

\subsection{Kronecker Product}

Kronecker product is a widely used matrix operation, and it is also the matrix representation of the tensor product in the standard basis. For matrices $\textbf{A}_{m\times n}$ and $\textbf{B}_{p \times q}$, their Kronecker product is a block matrix with size of $m\times p$ rows and $n \times q$ columns:
\begin{equation}
	\textbf{A}_{m\times n} \otimes \textbf{B}_{p \times q} = \begin{bmatrix} a_{1,1} \textbf{B}_{p \times q} & \cdots & a_{1,n} \textbf{B}_{p \times q} \\ \vdots  & \ddots & \vdots \\ a_{m,1} \textbf{B}_{p \times q} & \cdots & a_{m,n} \textbf{B}_{p \times q}  \end{bmatrix}
\end{equation}

\section{METHOD}
\label{sec:format}

\subsection{A Feature Fusion Method Based On Kronecker Product}
	In this section,  a Feature Fusion method based on Kronecker Product (KPFF) is propoesd. Given a group of features, let us denote them as $\textbf{x}_1, \textbf{x}_2,...,\textbf{x}_n$ (column vectors). Let us assume that the feature vectors are of the same dimension $r$, otherwise a single layer can be introduced to make the dimension of these vectors the same.

$\textbf{x}_i$ (i=1,2,...,n, the same below)  are multiplied by learnable weighted vectors $ \textbf{w}_i$ (column vectors with n dimensions) through Kronecker Product. Then the resulting vectors are added. The formula expression is:
\begin{equation}
	\textbf{y} = \sum_{i=1}^{n}  \textbf{w}_i \otimes 	\textbf{x}_i
\end{equation}

\begin{figure*}[htbp]
	\centering
	\includegraphics[width=180mm]{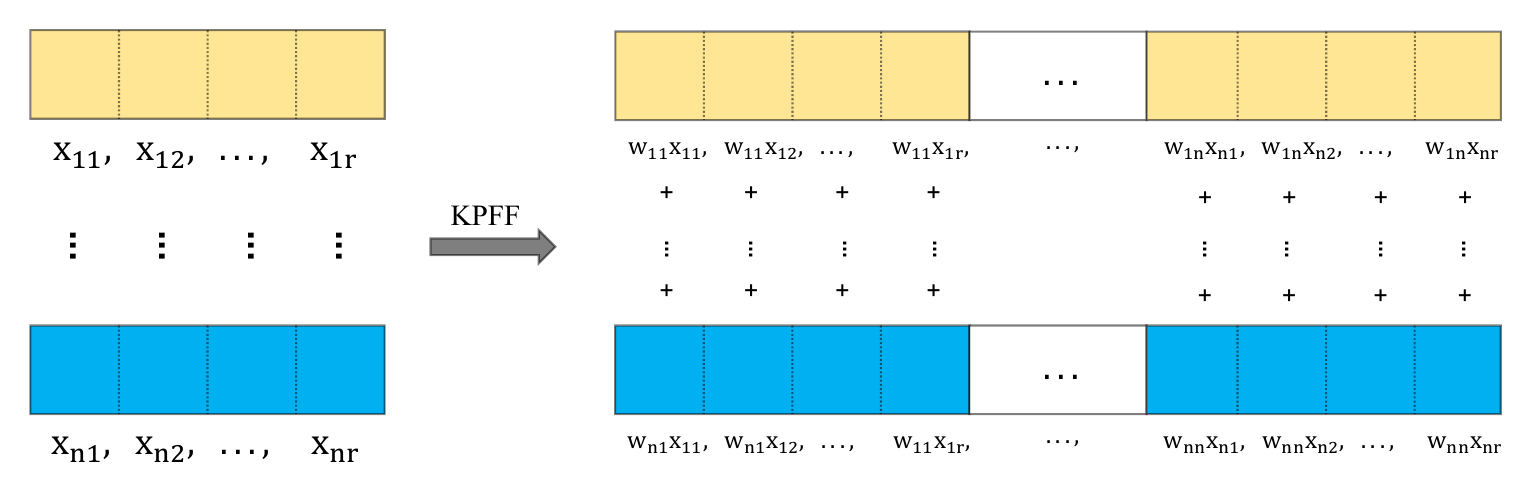}
	\caption{Kronecker Product Feature Fusion (KPFF)}
	\label{fig3:KPFF}
\end{figure*}

The column vector $\textbf{y}$ is the fusion result with $n\times r$ dimensions, and the learnable weighted vectors $\textbf{w}_1,\textbf{w}_2,...,\textbf{w}_n$ can be optimized through the Backpropagation procedure. 

In order to illustrate the relationship of KPFF and other two widely used feature fusion methods: add and concatenate, two special cases are discussed.

The first special case of KPFF is when $\textbf{w}_i$=$\textbf{e}_i$. $\textbf{e}_i$ represents the unit column vector where the i-th component is 1 and all other components are 0. Through the definition of Kronecker Product, it is easy to get that the outcome $\textbf{y}$ can be expressed as follows (where $||$  denotes the concatenate operation):
\begin{equation}
	\textbf{y} = ||_{i=1}^n	\textbf{x}_i
\end{equation}
Through the equation, this case can be regarded as the concatenate method.

The second special case of KPFF is when $\textbf{w}_i$=$\textbf{e}_1$. Similar to above-mentioned discussion, it is straightforward to get that the result $\textbf{y}$ can be expressed as follows (where $||$ denotes the concatenate operation, and $\textbf{0}_{(n-1) \times r}$ denotes the zero column vector with $(n-1) \times r$ dimensions):

\begin{equation}
	\textbf{y} =  (\sum_{i=1}^{n} 	\textbf{x}_i) || \textbf{0}_{(n-1) \times r}
\end{equation}

Considering that the zeros do not have influence on the follow-up layers (0 multiple any weight gets 0), this case can be deemed as the add method. In ResNet and Transformer, the fusion method of Add is employed, while in UNet, Feature Pyramid Network (FPN), and YOLO, the fusion method of Concat is adopted. These network architectures play crucial roles in their respective domains, highlighting the significance of feature fusion methods in convolutional neural networks.

In conclusion, KPFF becomes the add and the concatenate methods in two special cases.

\subsection{The Backpropagation Procedure Of KPFF}
In this section, the Backpropagation procedure of KPFF is discussed. Let us denote loss function as $L$, and the a-th component (a=1,2,...,n$\times$r) of $\textbf{y}$ as $y_a$,  the b-th component (b=1,2,...,n) of $\textbf{w}_i$ $(i=1,2,...,n)$ as $w_{i,b}$, and  the c-th component (c=1,2,...,r) of $\textbf{x}_j$ (j=1,2,...,n) as $x_{j,c}$. 

According to the chain rule, the partial derivative of $L$ with respect to $w_{i,b}$ can be expressed as:
\begin{equation}
	\frac{\partial L}{\partial w_{i,b}}= \sum_{a=1}^{n \times r} \frac{\partial L}{\partial y_{a}} \frac{\partial  y_{a}}{\partial w_{i,b}}
\end{equation}

The values of $ \frac{\partial L}{\partial y_{a}}$ have been calculated by the previous backpropagation procedure, and the values of $\frac{\partial  y_{a}}{\partial w_{i,b}}$ are required to be figured out.

Let us divide  the ${n \times r}$ coomponents into n groups, and then the a-th component is located in the $\lceil a/r \rceil $-th group. (where $\lceil \rceil $ is the symbol for rounding up) Since the components in the $\lceil a/r \rceil $-th group are only related to $ w_{i,b} $ when $b = \lceil a/r \rceil $, the values of $\frac{\partial  y_{a}}{\partial w_{i,b}}$ can be expressed as:
\begin{equation}
	\frac{\partial  y_{a}}{\partial w_{i,b}}= 
	\begin{cases}
		x_{i,(a- (\lceil a/r \rceil-1) \times r) } \quad b = \lceil a/r \rceil\\
		0 \quad others
	\end{cases}
\end{equation}

So the value of $\frac{\partial L}{\partial w_{i,b}}$ can be represented as:
\begin{equation}
	\frac{\partial L}{\partial w_{i,b}}= \sum_{a=(b-1) \times r+1}^{b \times r}   \frac{\partial L}{\partial y_{a}} x_{i,(a- (\lceil a/r \rceil-1) \times r) }
\end{equation}

Then the  weighted vectors $ \textbf{w}_i$ can be optimized according to the equation (10).

In addition, to complete the gradient calculation of previous layers, the patial derivative of $L$ with respect to $x_{j,c}$ should also be calculated. According to the chain rule: 
\begin{equation}
	\frac{\partial L}{\partial x_{j,c}}= \sum_{a=1}^{n \times r} \frac{\partial L}{\partial y_{a}} \frac{\partial  y_{a}}{\partial x_{j,c}}
\end{equation}

Similar to the situation of $ w_{i,b}$, the values of $\frac{\partial  y_{a}}{\partial  x_{j,c}}$ can be expressed as:
\begin{equation}
	\frac{\partial  y_{a}}{\partial x_{j,c}}= 
	\begin{cases}
		w_{j, \lceil a/r \rceil } \quad c = a- (\lceil a/r \rceil-1) \times r\\
		0 \quad others
	\end{cases}
\end{equation}

So the values of $\frac{\partial L}{\partial x_{j,c}}$ can be represented as:
\begin{equation}
	\frac{\partial L}{\partial x_{j,c}}= \sum_{k=1}^{n}   \frac{\partial L}{\partial y_{(k-1) \times r+c}}  	w_{j, k }
\end{equation}

Then the gradients of previous layers can be calculated  according to the equation (13).

\section{EXPERIMENTS}
\subsection{Datasets}
The UC-Merced LandUse dataset consists of images with a resolution of 256×256 pixels and a spatial resolution of 1 foot, all extracted from urban imagery data provided by the National Map of the United States. The dataset comprises 21 classes of land use, with each class having only 100 images. 

\subsection{Environment and settings}
 The hardware specifications are as follows: Processor: AMD Ryzen 9 5900HX with Radeon Graphics, octa-core; Memory: DDR4 3200MHz 16GB×2; Graphics Card: NVIDIA GeForce RTX 3080 Laptop GPU with 16GB VRAM. The experimental software environment is as follows: Python 3.8, PyTorch 1.12, and NumPy 1.23.5 under Windows 11.

In terms of parameter settings, the following parameters were selected through grid search: Adam optimizer was used with an initial learning rate of $1\times 10^{-4}$, weight decay was set to $5\times 10^{-4}$, BatchSize was set to 50, the maximum number of training epochs was set to 200, validation was conducted every 10 epochs, and the dropout probability was set to 0.5.

Regarding the experimental methodology, this paper employs a five-fold cross-validation approach. AlexNet, VGGNet, and Inception are utilized as backbone networks. A comparison is made between the original networks and the networks incorporating the KPFF method, evaluating the classification accuracy before and after the introduction of the module. Furthermore, for a comprehensive comparison, experiments were conducted using both the Add and Concat feature fusion methods.

This experiment did not utilize any pre-trained parameters; instead, the neural network was trained from scratch.

\subsection{Results and Analysis}
\begin{table}[t]\centering
	\caption{Five-Fold Cross-Validation for Evaluating Classification Accuracy of Original and Improved CNNs on the UCM Dataset}
	\begin{tabular}{cccc}
		\toprule
		\textbf{Method} & \textbf{AlexNet} & \textbf{VGGNet} & \textbf{Inception}\\
		\midrule
			original&88.10\% &89.52\%&88.57\% \\
	Add&90.71\%&91.19\%&89.76\% \\
	Concat&91.19\%&90.95\%&90.23\%\\
	\textbf{KPFF}&\textbf{91.90\%}&\textbf{92.62\%}&\textbf{92.14\%}\\
		\bottomrule
	\end{tabular}
	\label{methods}
	\vspace{-4mm}
\end{table}

From the experimental results, it can be observed that both Concat and Add methods enhance the classification performance of the original network. In this experiment, Add and Concat achieved higher classification accuracy in different scenarios. Meanwhile, the proposed KPFF method outperforms the original network as well as the Add and Concat methods in terms of classification performance. Compared to the original networks, the classification accuracy of KPFF-improved AlexNet, VGGNet, and Inception on the UCM dataset increased by 4.13\%, 3.46\%, and 4.03\%, respectively. The experimental results indicate that the KPFF feature fusion method effectively enhances CNN classification performance, and compared to Add and Concat methods, the proposed approach exhibits greater advantages.

\section{Conclusion}
\label{sec:typestyle}

This paper introduces a novel feature fusion method by incorporating the Kronecker product, extending two classic feature fusion methods, Add and Concat. Both theoretical analysis and experimental results demonstrate that the proposed approach is more effective in improving the accuracy of CNNs in remote sensing scene classification compared to Add and Concat methods.

The computational complexity of our method is comparable to that of the Concatenation method. However, exploring further reductions in computational complexity remains an area worthy of additional research. There are also new research directions in feature fusion, such as integrating feature fusion with coding theory or combining it with subspace learning, attention mechanisms, and so on. Investigating the relationships between these research methods and our approach, as well as whether they can further improve the accuracy of remote sensing image classification, is worth further exploration.

\bibliographystyle{IEEEtran}
\bibliography{ref}

\begin{thebibliography}{10}
\providecommand{\url}[1]{#1}
\csname url@samestyle\endcsname
\providecommand{\newblock}{\relax}
\providecommand{\bibinfo}[2]{#2}
\providecommand{\BIBentrySTDinterwordspacing}{\spaceskip=0pt\relax}
\providecommand{\BIBentryALTinterwordstretchfactor}{4}
\providecommand{\BIBentryALTinterwordspacing}{\spaceskip=\fontdimen2\font plus
\BIBentryALTinterwordstretchfactor\fontdimen3\font minus
  \fontdimen4\font\relax}
\providecommand{\BIBforeignlanguage}[2]{{%
\expandafter\ifx\csname l@#1\endcsname\relax
\typeout{** WARNING: IEEEtran.bst: No hyphenation pattern has been}%
\typeout{** loaded for the language `#1'. Using the pattern for}%
\typeout{** the default language instead.}%
\else
\language=\csname l@#1\endcsname
\fi
#2}}
\providecommand{\BIBdecl}{\relax}
\BIBdecl

\bibitem{review}
K.~Nogueira, O.~A. Penatti, and J.~A. Dos~Santos, ``Towards better exploiting
  convolutional neural networks for remote sensing scene classification,''
  \emph{Pattern Recognition}, vol.~61, pp. 539--556, 2017.

\bibitem{review2}
X.~Lu, X.~Zheng, and Y.~Yuan, ``Remote sensing scene classification by
  unsupervised representation learning,'' \emph{IEEE Transactions on Geoscience
  and Remote Sensing}, vol.~55, no.~9, pp. 5148--5157, 2017.

\bibitem{review3}
G.~Cheng, J.~Han, and X.~Lu, ``Remote sensing image scene classification:
  Benchmark and state of the art,'' \emph{Proceedings of the IEEE}, vol. 105,
  no.~10, pp. 1865--1883, 2017.

\bibitem{SIFT}
A.~Sedaghat, M.~Mokhtarzade, and H.~Ebadi, ``Uniform robust scale-invariant
  feature matching for optical remote sensing images,'' \emph{IEEE Transactions
  on Geoscience and Remote Sensing}, vol.~49, no.~11, pp. 4516--4527, 2011.

\bibitem{HOG}
C.~I. Patel, D.~Labana, S.~Pandya, K.~Modi, H.~Ghayvat, and M.~Awais,
  ``Histogram of oriented gradient-based fusion of features for human action
  recognition in action video sequences,'' \emph{Sensors}, vol.~20, no.~24, p.
  7299, 2020.

\bibitem{BOW}
X.~Chen, G.~Zhu, and M.~Liu, ``Bag-of-visual-words scene classifier for remote
  sensing image based on region covariance,'' \emph{IEEE Geoscience and Remote
  Sensing Letters}, vol.~19, pp. 1--5, 2022.

\bibitem{CH}
R.~M. Anwer, F.~S. Khan, and J.~Laaksonen, ``Compact deep color features for
  remote sensing scene classification,'' \emph{Neural Processing Letters},
  vol.~53, no.~2, pp. 1523--1544, 2021.

\bibitem{GLCM}
W.~Xia, L.~Yan, and H.~Xie, ``A dsm-based co-occurrence matrix for semantic
  classification,'' \emph{IEEE Geoscience and Remote Sensing Letters}, vol.~19,
  pp. 1--5, 2020.

\bibitem{CNN1}
T.~Kattenborn, J.~Leitloff, F.~Schiefer, and S.~Hinz, ``Review on convolutional
  neural networks (cnn) in vegetation remote sensing,'' \emph{ISPRS journal of
  photogrammetry and remote sensing}, vol. 173, pp. 24--49, 2021.

\bibitem{CNN2}
W.~Lee, D.~Sim, and S.-J. Oh, ``A cnn-based high-accuracy registration for
  remote sensing images,'' \emph{Remote Sensing}, vol.~13, no.~8, p. 1482,
  2021.

\bibitem{Alex}
X.~Han, Y.~Zhong, L.~Cao, and L.~Zhang, ``Pre-trained alexnet architecture with
  pyramid pooling and supervision for high spatial resolution remote sensing
  image scene classification,'' \emph{Remote Sensing}, vol.~9, no.~8, p. 848,
  2017.

\bibitem{VGG}
S.~Chaib, H.~Liu, Y.~Gu, and H.~Yao, ``Deep feature fusion for vhr remote
  sensing scene classification,'' \emph{IEEE Transactions on Geoscience and
  Remote Sensing}, vol.~55, no.~8, pp. 4775--4784, 2017.

\bibitem{Res}
G.~Cheng, J.~Han, and X.~Lu, ``Remote sensing image scene classification:
  Benchmark and state of the art,'' \emph{Proceedings of the IEEE}, vol. 105,
  no.~10, pp. 1865--1883, 2017.

\bibitem{Inception}
N.~Devi and B.~Borah, ``Refining the features transferred from pre-trained
  inception architecture for aerial scene classification,'' \emph{International
  Journal of Remote Sensing}, vol.~40, no.~24, pp. 9260--9278, 2019.

\bibitem{CNN3}
X.~Zhang, S.~Cheng, L.~Wang, and H.~Li, ``Asymmetric cross-attention
  hierarchical network based on cnn and transformer for bitemporal remote
  sensing images change detection,'' \emph{IEEE Transactions on Geoscience and
  Remote Sensing}, vol.~61, pp. 1--15, 2023.

\bibitem{CNN4}
J.~Shi, Y.~Wang, Z.~Yu, G.~Li, X.~Hong, F.~Wang, and Y.~Gong, ``Exploiting
  multi-scale parallel self-attention and local variation via dual-branch
  transformer-cnn structure for face super-resolution,'' \emph{IEEE
  Transactions on Multimedia}, 2023.

\bibitem{CNN5}
T.~Kattenborn, J.~Leitloff, F.~Schiefer, and S.~Hinz, ``Review on convolutional
  neural networks (cnn) in vegetation remote sensing,'' \emph{ISPRS journal of
  photogrammetry and remote sensing}, vol. 173, pp. 24--49, 2021.

\bibitem{FF}
Y.~Ding, Z.~Zhang, X.~Zhao, D.~Hong, W.~Cai, C.~Yu, N.~Yang, and W.~Cai,
  ``Multi-feature fusion: Graph neural network and cnn combining for
  hyperspectral image classification,'' \emph{Neurocomputing}, vol. 501, pp.
  246--257, 2022.

\bibitem{Ma}
C.~Shi, X.~Zhang, J.~Sun, and L.~Wang, ``Remote sensing scene image
  classification based on dense fusion of multi-level features,'' \emph{Remote
  Sensing}, vol.~13, no.~21, p. 4379, 2021.

\bibitem{Mei}
S.~Mei, K.~Yan, M.~Ma, X.~Chen, S.~Zhang, and Q.~Du, ``Remote sensing scene
  classification using sparse representation-based framework with deep feature
  fusion,'' \emph{IEEE Journal of Selected Topics in Applied Earth Observations
  and Remote Sensing}, vol.~14, pp. 5867--5878, 2021.

\bibitem{Xu}
K.~Xu, H.~Huang, Y.~Li, and G.~Shi, ``Multilayer feature fusion network for
  scene classification in remote sensing,'' \emph{IEEE Geoscience and Remote
  Sensing Letters}, vol.~17, no.~11, pp. 1894--1898, 2020.

\end{thebibliography}

\end{document}